\documentclass[10pt,twocolumn,letterpaper]{article}

\usepackage[pagenumbers]{cvpr} 

\usepackage{graphicx}
\usepackage{amsmath}
\usepackage[accsupp]{axessibility}
\usepackage{amssymb}
\usepackage{booktabs}
\usepackage[noend]{algpseudocode}

\usepackage{algorithmicx,algorithm}

\usepackage{times}
\usepackage{epsfig}
\usepackage{pifont}       
\usepackage{bbding}       
\usepackage{setspace}
\usepackage{multirow}
\usepackage{cite}
\usepackage{comment}
\usepackage{booktabs}
\usepackage{arydshln}
\usepackage{color}
\usepackage{xcolor}
\usepackage{bm}
\usepackage[normalem]{ulem}

\usepackage[pagebackref,breaklinks,colorlinks]{hyperref}

\usepackage[capitalize]{cleveref}
\crefname{section}{Sec.}{Secs.}
\Crefname{section}{Section}{Sections}
\Crefname{table}{Table}{Tables}
\crefname{table}{Tab.}{Tabs.}

\begin{document}

\title{CEAT: Continual Expansion and Absorption Transformer for Non-Exemplar Class-Incremental Learning}

\author{Xinyuan Gao\textsuperscript{\rm 1}, Songlin Dong\textsuperscript{\rm 2}, Yuhang He\textsuperscript{\rm 2}{\thanks{\* Corresponding authors}}, Xing Wei\textsuperscript{\rm 1}, Yihong Gong\textsuperscript{\rm 1}\\ 
\textsuperscript{\rm 1}School of Software Engineering, Xi'an Jiaotong University\\
\textsuperscript{\rm 2}Institute of Artificial Intelligence and Robotics, Xi’an Jiaotong University\\
{\tt\small \{gxy010317,dsl972731417\}@stu.xjtu.edu.cn} \\
{\tt\small heyuhang@xjtu.edu.cn}, {\tt\small \{weixing,ygong\}@mail.xjtu.edu.cn}}
\maketitle

\begin{abstract}
In real-world applications, dynamic scenarios require the models to possess the capability to learn new tasks continuously without forgetting the old knowledge. Experience-Replay methods store a subset of the old images for joint training. In the scenario of more strict privacy protection, storing the old images becomes infeasible, which leads to a more severe plasticity-stability dilemma and classifier bias. To meet the above challenges, we propose a new architecture, named continual expansion and absorption transformer~(CEAT). The model can learn the novel knowledge by extending the expanded-fusion layers in parallel with the frozen previous parameters. After the task ends, we losslessly absorb the extended parameters into the backbone to ensure that the number of parameters remains constant. To improve the learning ability of the model, we designed a novel prototype contrastive loss to reduce the overlap between old and new classes in the feature space. Besides, to address the classifier bias towards the new classes, we propose a novel approach to generate the pseudo-features to correct the classifier. We experiment with our methods on three standard Non-Exemplar Class-Incremental Learning~(NECIL) benchmarks. Extensive experiments demonstrate that our model gets a significant improvement compared with the previous works and achieves 5.38\%, 5.20\%, and 4.92\% improvement on CIFAR-100, TinyImageNet, and ImageNet-Subset.
\end{abstract}


\begin{figure*}[t]
\begin{center}
    \includegraphics[width=0.9\textwidth]{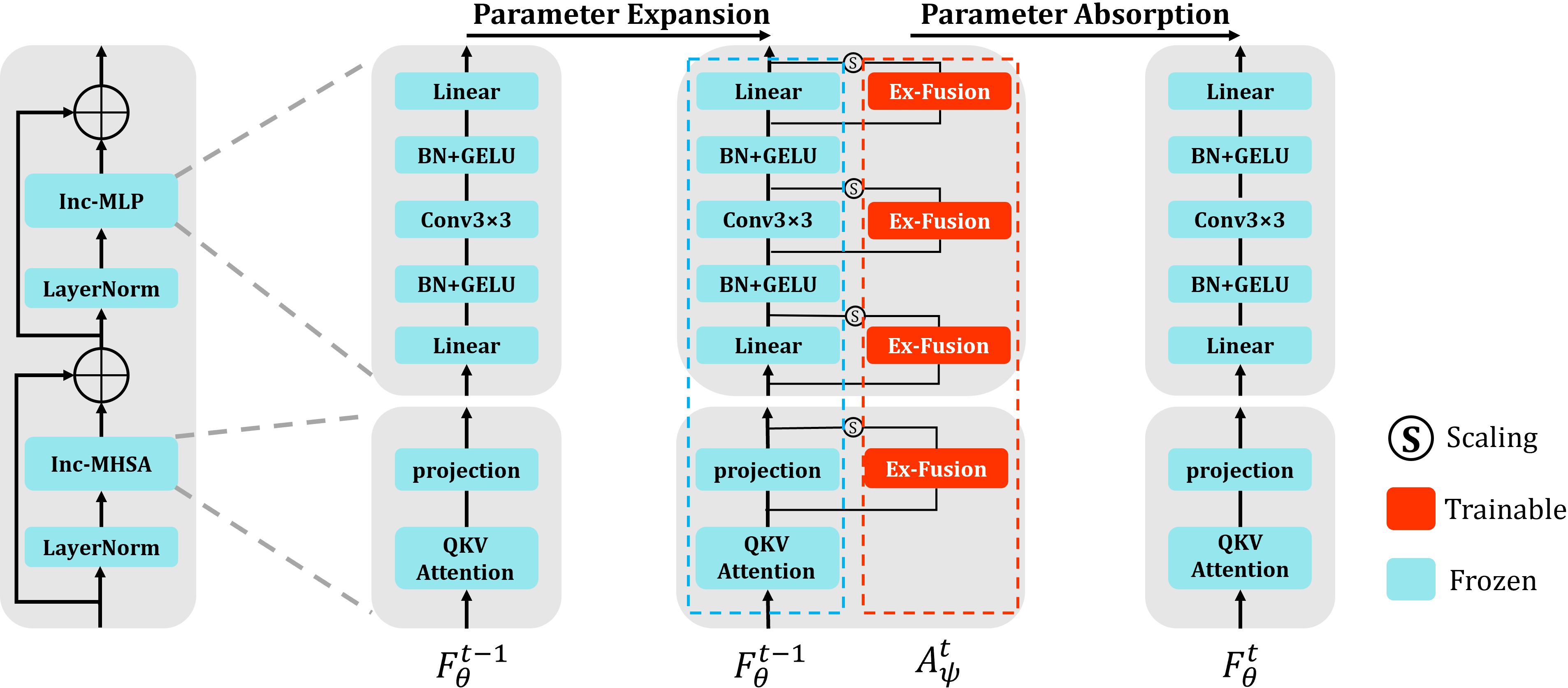}
\end{center}
\vspace{-0.2cm}
\caption{The overall process of our continual expansion and absorption. Each ViT layer consists of two LayerNorm layers, one incremental MHSA (Inc-MHSA), and one incremental MLP (Inc-MLP). In task $t > 0$, the backbone $F_{\theta}^{t-1}$ is frozen, and the ex-fusion parameters $A_{\psi}^{t}$ are trainable to learn the new task. After the task ends, parameters $A_{\psi}^{t}$ are absorbed losslessly into the backbone for the next task.}
\label{fig:Adaframework}
\vspace{-0.3cm}
\end{figure*}

\section{Introduction}
Class Incremental Learning~(CIL) aims to recognize new classes emerging from incrementally obtained data without suffering from catastrophic forgetting\cite{CF}-performance disastrous deterioration on the previously trained data. Some CIL methods\cite{lucir,ICARL,der,dytox,tao2020topology} store a subset of previous exemplars in the memory buffer and replay them in the new tasks. However, in many real-world applications, the storage of previous exemplars is restricted due to privacy and security concerns. Therefore, we focus on a more practical and challenging setting of \textbf{Non-Exemplar Class Incremental Learning~(NECIL)}.

Different from the general CIL setting, NECIL does not allow the model to store any old image samples and prohibits the use of any pre-trained models\cite{EWC,LWF,yu2020semantic,shi2023prototype,petit2023fetril,zhu2021prototype,zhu2022self}. In recent works, \cite{zhu2021prototype} uses the augmenting prototypes of old classes to maintain the decision boundary and employ self-supervised learning to learn more generalizable and transferable features. \cite{zhu2022self} proposes a dynamic structure reorganization strategy to update the feature space and a prototype selection mechanism to reduce the feature confusion among similar classes.

Despite these methods undertaken numerous efforts to meet the challenge of NECIL, two primary challenges persist: \textit{plasticity-stability dilemma}\cite{carpenter1987massively}, which indicates that a model needs to be stable to maintain the old knowledge (\emph{i.e.} stability) and be plastic to learn new one (\emph{i.e.}, plasticity), and \textit{classifier bias}\cite{ICARL}, which indicates that a model’s classifier is biased towards the newly emerged classes and is more likely to categorize objects into the new classes.

To address the above problems, we propose a novel architecture, named \textbf{Continual Expansion and Absorption Transformer~(CEAT)} for the NECIL problem without the large-scale pre-trained model. Our approach contains two major steps, \emph{i.e.}, parameter expansion and parameter absorption. Specifically, as shown in Fig~\ref{fig:Adaframework}, at each incremental learning task $t$, we construct a network with a frozen backbone to be stable of old knowledge and a set of trainable layers to be plastic for learning new ones.  We expand a trainable \textit{expanded-fusion layer~(ex-fusion layer)} parallelly for some parameterized operation layer of the backbone, including convolution and linear projection in the MLP block, and linear projection in the multi-head attention block. Only the ex-fusion layers are optimized while the original backbone is kept frozen during the training phase except for the first task. Subsequently, after the training of each session $t$, the optimized ex-fusion layers are \emph{absorbed} into the backbone network by parameter-weighted summing. This remains the network structure and parameter number of the model unchanged after increment learning tasks, which is implement-friendly for practical usages and resource-constrained edge devices. To improve the ability to learn new knowledge, we propose a new loss function, named \textit{prototype contrastive loss}~(PCL), which can achieve inter-class separation and reduce the overlap between the old and new classes. Moreover, \textit{to alleviate the classifier bias to the new classes}, we design a novel mechanism to generate the pseudo-features to correct the bias of classifiers and dynamically maintain the decision boundary of previous classes. We conduct extensive experiments on the well-established CIFAR-100, TintImageNet and ImageNet-Subset benchmarks and boast significant performance gains.

The main \textbf{contribution} of this paper is summarized as follows: 
\begin{itemize}
\vspace{-0.2cm}
\item We propose a novel ViT architecture called CEAT, to solve the NECIL problems. CEAT optimizes the ex-fusion layers while freezing the previous parameters and absorbs the expanded parameters losslessly after the task ends.
\vspace{-0.2cm}
\item We introduce a novel approach for generating pseudo-features that allows the model to dynamically maintain the decision boundary of previous classes.
\vspace{-0.2cm}
\item We develop a prototype contrastive loss to achieve inter-class separation and reduce the overlapping among the classes.
\vspace{-0.2cm}
\item We conduct comprehensive experiments on three commonly used image classification benchmarks. The experimental results demonstrate that our proposed method outperforms previous approaches.
\end{itemize}

\begin{figure*}[t]
\begin{center}
    \includegraphics[width=0.9\textwidth]{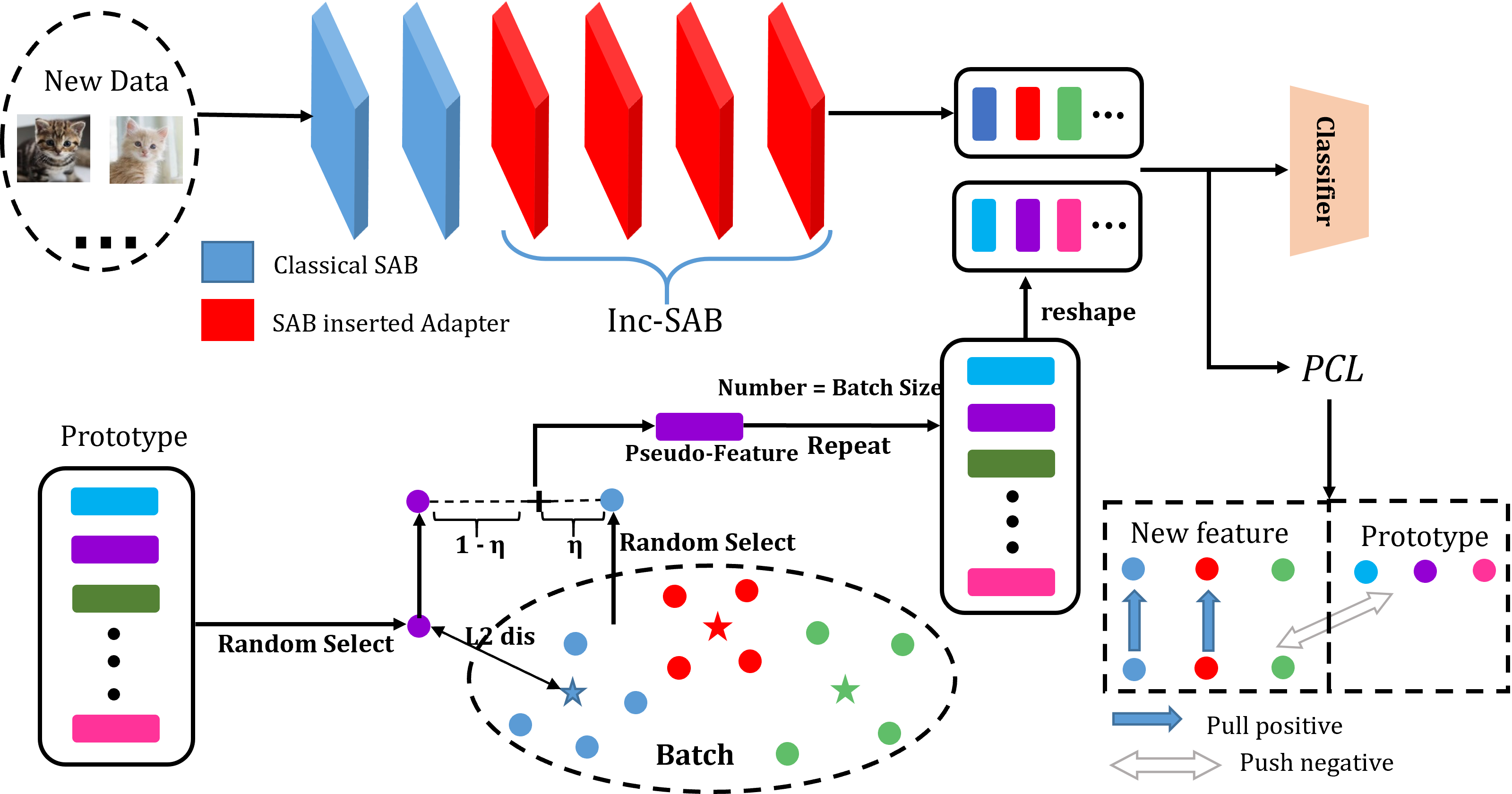}
\end{center}
\caption{The overall structure of our model. Our model consists of two classical SABs~(Self-Attention Blocks), four Inc-SABs, which have expanded parameters, and a classifier. The current images pass the feature extractor and the resulting features concat with the pseudo-features to balance the classifier. Also, the concat features are used to compute the PCL~(prototype contrastive loss) to reduce the overlap among the old and new classes.}
\label{fig:framework}
\end{figure*}

\section{Related Work}
Among the domains of IL, Class Incremental Learning (CIL) is a more challenging problem and has greater practical value, which has resulted in its recent surge in attention in research. Depending on whether samples need to be stored, CIL can be divided into two types: Experience-Replay CIL and Non-Exemplar CIL.

\vspace{-0.2cm}

\paragraph{Experience-Replay Class Incremental Learning.} Previous works\cite{tao2020topology,der,dytox,wang2022foster,gao2023dkt,ICARL,lucir} retain the exemplars from previous tasks in a memory buffer to replay in the new task. Current experience-replay methods can be divided into the following two classes. \textit{Knowledge distillation methods}\cite{ICARL,lucir,tao2020topology,dytox,podnet,kang2022class} limit the output difference between the old and new networks on the data of the current task. \cite{lucir,ICARL,dytox} force the new network to reproduce similar logits to the old network to reduce forgetting. \cite{podnet,kang2022class} align the intermediate feature outputs of the old and new networks to improve the performance. \cite{tao2020topology,dong2021few,topic} use old class data to maintain the topology structure of old feature space to alleviate forgetting. \textit{Dynamic architecture methods}\cite{der,wang2022foster,zhou2022model} expand extra feature extractors for each task and concatenates the feature from each feature extractor for classification. However these methods typically require more training memory and computing resources.

Nevertheless, on account of considerations related to privacy, security, and legal regulations, private user exemplars cannot be stored in many important scenarios of real-world applications. At the same time, storing old image samples requires a lot of additional memory overhead. These motivate us to focus on the non-exemplar setting, which is more privacy-protected.

\vspace{-0.2cm}

\paragraph{Non-Exemplar Class Incremental Learning.} Recent works\cite{LWF,yu2020semantic,zhu2021prototype,zhu2022self,shi2023prototype} focus on the more challenging but more practically valuable problem of NECIL without the pre-trained model. Since NECIL does not store any old image exemplars except for a prototype per class, it encounters increased difficulty in dealing with the plasticity-stability dilemma and addressing classifier bias. Previous methods all focus on how to overcome these two problems. PASS\cite{zhu2021prototype} leverages augmenting prototypes of old classes to preserve the decision boundary and employs self-supervised learning to acquire more generalizable and transferable features. SSRE\cite{zhu2022self} proposes a dynamic structure reorganization strategy to update the feature space and a prototype selection mechanism to reduce the feature confusion among similar classes. FeTrIL\cite{petit2023fetril} freezes the feature extractors after the first task and uses a simple pseudo-feature generator to maintain the decision boundary. These three methods are our main comparison objects.

\vspace{-0.2cm}

\paragraph{Class Incremental Learning in Vision Transformer.} Recent works\cite{dytox,gao2023dkt,l2p,dualprompt,smith2023coda} have introduced the ViT architecture into the continual learning field. \cite{l2p,dualprompt,smith2023coda} introduce the large-scale pre-trained model into the continual learning field. They freeze the pre-trained model~(ViT B16 pre-trained on ImageNet21K) as the backbone and optimize the prompts to tune the model for the downstream tasks. L2P\cite{l2p} proposes a key-query matching strategy to select the prompts from the prompt pool, and then insert them into the pre-trained model. CODA\cite{smith2023coda} weighting the prompts and can be optimized in an end-to-end fashion. However, These methods heavily depend on a large-scale pre-trained model. Thus these methods are difficult to deploy in resource-constrained edge devices due to the large pre-trained model. Moreover, when facing a new dataset that has a significant distance from the pre-trained data, these methods often perform satisfactorily.
\textit{In this paper, we do not compare with such methods because their performance declines disastrously without the large-scale pre-trained model}. 

Some methods\cite{dytox,gao2023dkt,hu2023dense} are experience-replay methods without the pre-trained model. DyTox\cite{dytox} and DKT\cite{gao2023dkt} design various tokens to retain the information from previous tasks and use the old samples to correct the classifier bias. DNE\cite{hu2023dense} propose a network-expanded method to reduce the forgetting and balance the trade-off between accuracy and model size.

However, \cite{yu2021improving} have showcased that ViTs suffer from more serious forgetting when trained from scratch. Thus applying ViT architecture to the NECIL problem is still lacking research. To meet this challenge,  we attempt to propose a novel architecture based on the vision transformer, which has a similar parameter amount to previous works and can be trained from scratch, for the NECIL problem.

\section{Methodology}
\subsection{Problem Setting}
Given a dataset $D=\{D_t\}_{t=0}^T$ of $T$ incremental learning sessions, where $D_0$ is used to train a base model and $D_t~(t>0)$ is used for incremental learning. Each element $D_t=\{X_t,Z_t\}$ of $D$ is composed of a training set $X_t$ and a testing set $Z_t$. For each training set $X_t=\{(x_t^i,y_t^i)\}_{i=1}^{N_t}$, $x_t^i$ denotes the $i$-th sample, $y_t^i\in C_t$ denotes the class label, where $C_t$ denotes the label set of the $t$-th session. The objective of NECIL is to incrementally train a unified DNN model across the $T$ sessions without requiring storing old class samples during the learning process. Specifically, at each session $t$, only $X_t$ is available for training, and the model is evaluated on a combined testing set $Z_{0 \sim t} = Z_0 \cup \cdots \cup Z_t$ and is required to recognize all the old classes encountered $C_{0 \sim t} = C_0 \cup \cdots \cup C_t$.

\subsection{Continual Expansion and Absorption}
For simplicity, we define the feature extractor as $F^t(\cdot, \theta)$, the classifier as $g(\cdot, \phi)$, and the ex-fusion layer as $A^t(\cdot, \psi)$, where t is the task id. For the first phase, we optimize the model $F_\theta^0$ on dataset $D_0$. After the first phase, we repeatedly execute our expansion and absorption operation for incremental learning.


\paragraph{Parameter Expansion.} At each task $t$, we first freeze the feature extractor $F_\theta^{t-1}$ from the previous task to maintain the stability of the model. The classifier $g_\phi^{t-1}$ extends the head to $g_\phi^{t}$, which is trainable, to accommodate the increase of categories. To learn the new task, we define a series of trainable parameters of ex-fusion layer $A_\psi^t$ = $A_{\psi_1}^t \cdots A_{\psi_i}^t$, where $i$ is the index of the ex-fusion layer. To adhere to the principles of simplicity and effectiveness, the ex-fusion layer is defined solely as a bias-free linear layer
\begin{equation}
\begin{aligned}
A_{\psi_i}^t(x_i) = \psi_i * x_i
\end{aligned}
\end{equation}
where $\psi_i$ is the weight parameters of the ex-fusion layer and $x_i$ is the input feature map.

During the incremental training phase, the $A_\psi^t$ are extended in parallel with the different frozen backbone module $F_{\theta_1}^{t-1} \cdots F_{\theta_i}^{t-1}$. We optimize the $A_\psi^t$ to force the model to learn new knowledge. The input feature map $x_i$ is fed into the expanded layer and the backbone module respectively. Then the results are summed into a new feature map to the next layer

\begin{equation}
\begin{aligned}
x_{i+1} = F_{\theta_i}^{t-1}(x_i) + \lambda * A_{\psi_i}^{t}(x_i)
\end{aligned}
\end{equation}
where $\lambda$ is the scaling parameter in different incremental scenarios. $\lambda=Num(C_t)/10$, where $Num(C_t)$ is the number of $C_t$. The following $\lambda$ has the same meaning as this equation.

\paragraph{Parameters Absorption.} To avoid the continuous augmentation in both memory usage and the number of parameters, we absorb the extended parameters $A_\psi^t$ into the main backbone $F_\theta^{t-1}$ to form a new backbone upon task completion
\begin{equation}
\begin{aligned}
F^t_\theta = F^{t-1}_\theta + \lambda * A^t_\psi \\
\end{aligned}
\end{equation}

After completion of task t, all the expanded parameters $A^t_\psi$ are absorbed losslessly into the main backbone to maintain a constant total parameter amount. This significantly enhances the practicality of our model in different incremental scenarios.

\paragraph{Absorption implementation.} The classical vision transformer model have three typical trainable modules in each block: the MHSA~(Multi-Head Self-Attention) module, the linear layer, and the convolutional layer. To highlight the comprehensive consideration, we discuss the implementation of lossless absorption in the above three modules, respectively. 

For the MHSA module, which consists of a QKV-Attention and a projection layer. We opt to expand the parameters in parallel with the projection layer to enable dynamic adaptation, instead of the QKV-Attention. The reason is that despite the QKV-Attention module varies across different ViT blocks, all of them possess a projection layer in the MHSA. Without loss of generality, we show the MHSA of $l$-th layer. The detail is as follows:
\begin{equation}
\begin{aligned}
&\text{$x_a$} = \text{Softmax}\left(\frac{Q_l \cdot K_l^T}{\sqrt{d}}\right) V_l, \\
&x_{l1} = \text{Proj}(\text{$x_a$}), \quad x_{l2} = \lambda \cdot A_\psi(\text{$x_a$}), \\
&x_l = x_{l1} + x_{l2}.
\end{aligned}
\end{equation}
where $Proj$ represents a projection layer and $Q_l$. $K_l$, $V_l$ represents the standard QKV operations in the vision transformer.

After the training phase, since the projection layer is essentially a linear layer, the lossless absorption is equivalent to the absorption of the linear layer in the following.

For the linear layer, we can simply sum the two linear layer weights to fuse two linear layers into a single linear layer
\begin{equation}
\begin{aligned}
&W_{new} = W_{linear} + \lambda * \psi, \\
\end{aligned}
\end{equation}
where $W_{linear}$ is the linear layer weight in the main backbone and $W_{new}$ is the linear weight after absorption. 

As both the $1\times1$ convolution and linear layers are consistent in structure and function, it is easy to implement by zero-padding the $1\times1$ kernel to $3\times 3$ and adding the two kernels up 
\begin{equation}
\begin{aligned}
&W_{3*3} = ZP(\psi), \\
&W_{new} = W_{1} + \lambda * W_{3*3}
\end{aligned}
\end{equation}
where ZP represents the zero-padding operation, $W_1$ represents the $3\times 3$ convolution in the main backbone.

\subsection{Batch Interpolation Pseudo-Features}
\paragraph{Prototype.} Due to the risks of data leakage and privacy violations, the NECIL problem does not permit the storage of any past samples except for one prototype per class. Following the recent research, we calculate the class center vector for each class and maintain it as the prototype
\begin{equation}\label{eq:angle}
\begin{aligned}
P_{t,k} = \frac{1}{N_{t,k}}\sum_{n=1}^{N_{t,k}} F_\theta(x_k^t)
\end{aligned}
\end{equation} 
where t is a new task, $x_k^t$ is an image of class k in task t, and $N_{t, k}$ is the number of training images of class k. We set the prototypes P = $\{P_1, ..., P_i, ...\}$. The number of prototypes is equal to the number of old classes.


To preserve the decision boundaries and correct the classifier, we propose batch interpolation to generate pseudo-features. Consistent with recent research, we only generate an equal number of pseudo-features for each batch to ensure a fair comparison.

\paragraph{Feature Select.} Firstly  We calculate the class center of each new class in the batch. We define the class center S = $\{S_1, ..., S_j, ...\}$ in the batch of new training data
\begin{equation}\label{eq:angle}
\begin{aligned}
S_j = \frac{1}{M_{t,j}}\sum_{i=1}^{M_{t,j}} F_\theta(x_j^t)
\end{aligned}
\end{equation} 
where M is the number of features of this class in the batch. Then we randomly select a prototype $P_i$ to generate the pseudo-features. We compute the distance between $P_i$ and S in the feature space by L2 distance and select the nearest class center from S.

\paragraph{Interpolation.} We assume that the nearest class center vector is $S_j$, which belongs to the class $j$. We use interpolation to generate the pseudo-features to maintain the decision boundary. without loss of robustness, we randomly select a feature that belongs to the class $j$, named $F_j$. We use a similar approach to interpolation
\begin{equation}\label{eq:mixup}
\begin{aligned}
F_{ipf} = (1 - \eta) P_i + \eta F_j
\end{aligned}
\end{equation} 
the $F_{ipf}$ has the same label as $P_k$. $\eta$ is a hyperparameter to control the position of the pseudo feature. The $\eta\sim\zeta*Beta(0.8,0.5)$, where $\zeta$ increases linearly with tasks from 0.5 to 0.7.

The benefit of using interpolation to generate the pseudo-features is two-fold. Firstly, the pseudo feature is generated by the nearest class and the prototype. Hence the decision boundary of the old classes is flexible, which leads to a better pseudo-feature distribution. Secondly, the pseudo-features mix information from old and new classes and dynamically adjust with the batch. Therefore, batch interpolation pseudo-feature effectively solves the problem that the decision boundary maintained by Gaussian noise gradually loses efficacy as tasks increase. As the feature space keeps changing, the decision boundaries can still effectively distinguish the old and new classes.

\subsection{Prototype Contrastive Loss}
Previous methods usually ignore the importance of inter-class separation in the NECIL problem. To tackle the problem and enhance the ability to learn, we propose a prototype contrastive loss~(PCL) scheme. Under the unavailability of old class data, prototype contrastive learning achieves inter-class separation and reduces the overlap between old and new classes.

During the $t$-th training phase, our model stores a series of frozen prototypes P = $\{P_1, ..., P_i, ... ,\}$. The feature extractor $F_\theta$ projects an image x to a feature $z = F_\theta(x)$. The classifier $g_\phi$ projects the feature $z$ to a vector $\hat{y} = g_\phi(z)$. Without training the $g_\phi$, given a batch of training images B = $\{(x_i, y_i)\}_{i=1}^N$. The feature map is trained to minimize the prototype contrastive loss

\begin{equation}
\begin{aligned}
L_{pcl} = \sum_{i\in B}\frac{-1}{|p_i|}\sum_{p\in{p_i}}log\left(\frac{exp(z_i\cdot z_p / \tau)}{\sum_{z_j\in A(i)}exp(z_i\cdot z_j / \tau)}\right) \\
+ \sum_{i\in B\bigcup P}\frac{-1}{|p_i|}\sum_{p\in{p_i}}log\left(\frac{exp(z_i\cdot z_p / \tau)}{\sum_{z_j\in A(i)\bigcup P}exp(z_i\cdot z_j / \tau)}\right)
\end{aligned}
\end{equation}

where $\tau$ is a temperature hyperparameter; $A(i)$ is the set of the feature in the augmented batch $B$. $p_i = \{p \in A(i): \hat{y_p} = \hat{y_i}\}$ is the index set of positive samples, which have the same label.

The prototypes serve as an anchor to force new class features away from the feature distribution of old classes to maintain the distribution. Moreover, the PCL achieves inter-class separation between old and new classes instead of the mere separation among new classes in the previous contrastive loss. This improvement alleviates the confusion between old and new classes, which enhances the learning ability of new classes without forgetting more old knowledge. The ablation experiments demonstrate the effectiveness of prototype contrastive loss.

\begin{table*}[ht]
  \renewcommand{\arraystretch}{1.2}
  \renewcommand{\tabcolsep}{8.pt}
  \centering
  \resizebox{\linewidth}{!}{
    \begin{tabular}{l|l|ccc|ccc|c}
      \hline
      \multicolumn{1}{c|}{} & \multicolumn{1}{c|}{} & \multicolumn{3}{c|}{\textbf{CIFAR-100}}   & \multicolumn{3}{c|}{\textbf{TinyImageNet}} & \multicolumn{1}{c}{\textbf{ImageNet-Subset}} \\ \cline{3-5} \cline{6-8} \cline{9-9} 
      \multicolumn{1}{c|}{\multirow{-2}{*}{\textbf{Methods}}} & \multicolumn{1}{c|}{\multirow{-2}{*}{\textbf{Buffer size}}} & \textit{P=5} & \textit{P=10} & \textit{P=20} & \textit{P=5} & \textit{P=10} & \textit{P=20} & \textit{P=10}\\
      \hline
      iCaRL-CNN\cite{ICARL} &\multirow{5}{*}{20/class} & 51.07  & 48.66 & 44.43 & 34.64 & 31.15 & 27.90 & 50.53\\
      iCaRL-NCM\cite{ICARL} &  & 58.56 & 54.19 & 50.51  & 45.86 & 43.29 & 38.04 & 60.79  \\ 
      EEIL\cite{EEIL} &  & 60.37 & 56.05 & 52.34  & 47.12 & 45.01 & 40.50 & 63.34 \\
      UCIR\cite{lucir} & & 63.78 & 62.39 & 59.07  & 49.15 & 48.52 & 42.83 & 66.16 \\ 
      DyTox-ViT\cite{dytox}  &  &69.84 &67.68 & 63.23 &58.16 & 57.69& 54.74&74.43\\
      \hline
      EWC\cite{EWC} &\multirow{8}{*}{0/class} & 24.48 & 21.20 & 15.89 & 18.80 & 15.77 & 12.39 & 20.40 \\ 
      LwF\cite{LWF} &  & 45.93 & 27.43 & 20.07 & 29.12 & 23.10 & 17.43 & 31.18  \\
      MUC\cite{liu2020more} &  & 49.42 & 30.19 & 21.27& 32.58 & 26.61 & 21.95 & 35.07 \\ 
      SDC\cite{yu2020semantic} &  & 56.77 & 57.00 & 58.90 &- & - & - & 61.12  \\ 
      PASS\cite{zhu2021prototype} &  & 63.47 & 61.84 & 58.09  & 49.55 & 47.29 & 42.07 & 61.80\\
      SSRE\cite{zhu2022self} &  & 65.88 & 65.04 & 61.70 & 50.39 & 48.93 & 48.17 & 67.69  \\
      FeTrIL\cite{petit2023fetril} & &  66.30 &65.20 &61.50  &54.80 &53.10 &52.20 &71.20\\
      CEAT &  & \textbf{71.68}\footnotesize{\color{red} \textbf{+5.38}} & \textbf{69.46}\footnotesize{\color{red} \textbf{+4.26}} & \textbf{65.14}\footnotesize{\color{red} \textbf{+3.44}} &\textbf{59.89}\footnotesize{\color{red} \textbf{+5.19}} & \textbf{58.30}\footnotesize{\color{red} \textbf{+5.20}} & \textbf{55.89}\footnotesize{\color{red} \textbf{+3.79}}  & \textbf{76.12}\footnotesize{\color{red} \textbf{+4.92}} \\
      \hline
    \end{tabular}
  }
  \caption{Comparisons of the average incremental accuracy (\%) at different settings on CIFAR-100, TinyImageNet, ImageNet-Subset. P represents the number of phases, and buffer size represents the number of exemplars. The results are obtained from SSRE\cite{zhu2022self} and FeTrIL\cite{petit2023fetril}. We reproduced the ViT-based method DyTox\cite{dytox} of experience-replay CIL  for further comparison. The improvement compared to the SOTA results is in the last row.}
  \label{tab:sota}
\vspace{-0.2cm}
\end{table*}

\subsection{Optimization Objective}
As the ex-fusion layers continuously modify the feature space of the model, we require distillation to constrain the changes made to the feature space. Following the previous works, we employ the knowledge distillation (KD)\cite{KD}, including the logit distillation $L_{ld}$ and feature distillation $L_{fd}$, to constrain the feature extractor
\begin{equation}
\begin{aligned}
L_{kd} = L_{ld} +  L_{fd}
\end{aligned}
\end{equation}
where $L_{fd}$ uses the $L_1$ distance. By combining the prototype contrastive loss $L_{pcl}$ and interpolation pseudo-feature loss $L_{ipf}$ that we proposed above, we can obtain the comprehensive loss function $L_{total}$
\begin{equation}
\begin{aligned}
L_{total} =  L_{bce} + \alpha L_{kd} + \mu L_{ipf} +\delta L_{pcl}
\end{aligned}
\end{equation}
where $L_{bce}$ is used to train the new classes. $\delta$, $\alpha$, and $\mu$ are the hyperparameters. $\delta$ is set to 0.5. We set $\alpha= Num(C_{1:t})/Num(C_t)$ as done by previous work\cite{dytox,gao2023dkt}, and $\mu= Num(C_{1:t}) / 20$, which removes the need to finely tune this hyperparameter in different datasets. The details are added in the Appendix.

\section{Experiment}
\subsection{Experimental Settings}
\paragraph{Datasets.} We use three common image classification benchmarks: CIFAR100\cite{cifar}, TinyImageNet\cite{le2015tiny} and ImageNet-Subset\cite{deng2009imagenet} following recent work\cite{zhu2021prototype,zhu2022self,petit2023fetril}. CIFAR-100~\cite{cifar} contains 100 classes, each class containing 500 training images and 100 test images. Each image is represented by 32 $\times$ 32 pixels, which is particularly small for ViT architecture. Tiny-ImageNet has 200 classes, each class containing 500 training images, 50 testing images, and 50 validation images. Each image is represented by $64 \times 64$.  ImageNet-Subset is a 100-class subset of ImageNet~\cite{ImageNet}. ImageNet-Subset contains about 130000 images for training and 50 images per class for testing. Each image is represented by $224 \times 224$.

\paragraph{Experimental Setup.} We design three standard NECIL settings like previous works\cite{zhu2021prototype,zhu2022self,petit2023fetril} to validate the incremental performance of our model. For CIFAR100, we set 50 classes as the base training data and increased 5/10 classes per task. We also set 40 classes as the base training data and increased 3 classes per task to test long-range increment learning ability. For TinyImageNet, we set 100 classes as the base training data and increased 5/10/20 classes per task. For ImageNet-Subset, we set 50 classes as the base training data and increased 5 classes per task. The random seed is 1993 following the previous work.

\paragraph{Evaluation Metric.} Following the ~\cite{zhu2021prototype,zhu2022self,petit2023fetril}, we employed the average incremental accuracy~\cite{ICARL} and average forgetting~\cite{chaudhry2018riemannian} as the main evaluation metric. The average incremental accuracy is calculated by averaging the accuracy of each task, serving as a crucial metric for assessing overall incremental performance. The average forgetting is calculated on average by the difference between the accuracy rate after each task and the final accuracy rate, representing the model's ability to mitigate catastrophic forgetting.

\paragraph{Implementation Details.} The ViT-based feature extractor $F(\cdot,\theta)$ of our model consists of six SABs like DyTox\cite{dytox} to satisfy the limit of parameters in the NECIL problem. To compare with other methods that use ResNet18\cite{he2016deep} as the backbone fairly, the final number of parameters of our model is 10.89M in CIFAR100 and 11.10M in ImageNet-Subset, slightly less than ResNet18~(11.22M).  More training information will be provided in the Appendix.

\subsection{Main Properties and Analysis}

\paragraph{Average Accuracy.} We compare our method with other methods based on the average accuracy and report it in Table \ref{tab:sota}. The experimental results indicate that our method achieves significantly better performance compared to the previous NECIL method\cite{zhu2022self,petit2023fetril}. In CIFAR100, our method surpasses the second-best by up to \textbf{5.38\%}, \textbf{4.26\%}, and \textbf{3.44\%}, which demonstrates the ability to adapt the transformer architecture to the NECIL problem on small datasets. In TinyImageNet and ImageNet-Subset, our method surpasses the second best by up to \textbf{5.19\%}, \textbf{5.20\%}, and \textbf{3.79\%} on TinyImageNet and \textbf{4.92\%} on ImageNet-Subset. Moreover, It is worth noting that our method exhibits remarkable improvements over the classical example-based approaches, further alleviating the dependence of incremental learning on examples. 

\paragraph{Average Forgetting.} We report the average forgetting of different methods in Table \ref{tab:forget}. Compared to other state-of-the-art methods\cite{zhu2021prototype,zhu2022self,petit2023fetril}, our approach demonstrates a notable advantage in terms of average forgetting across CIFAR-100 and TinyImageNet.

\begin{figure}[t]
\begin{center}
    \includegraphics[width=0.45\textwidth]{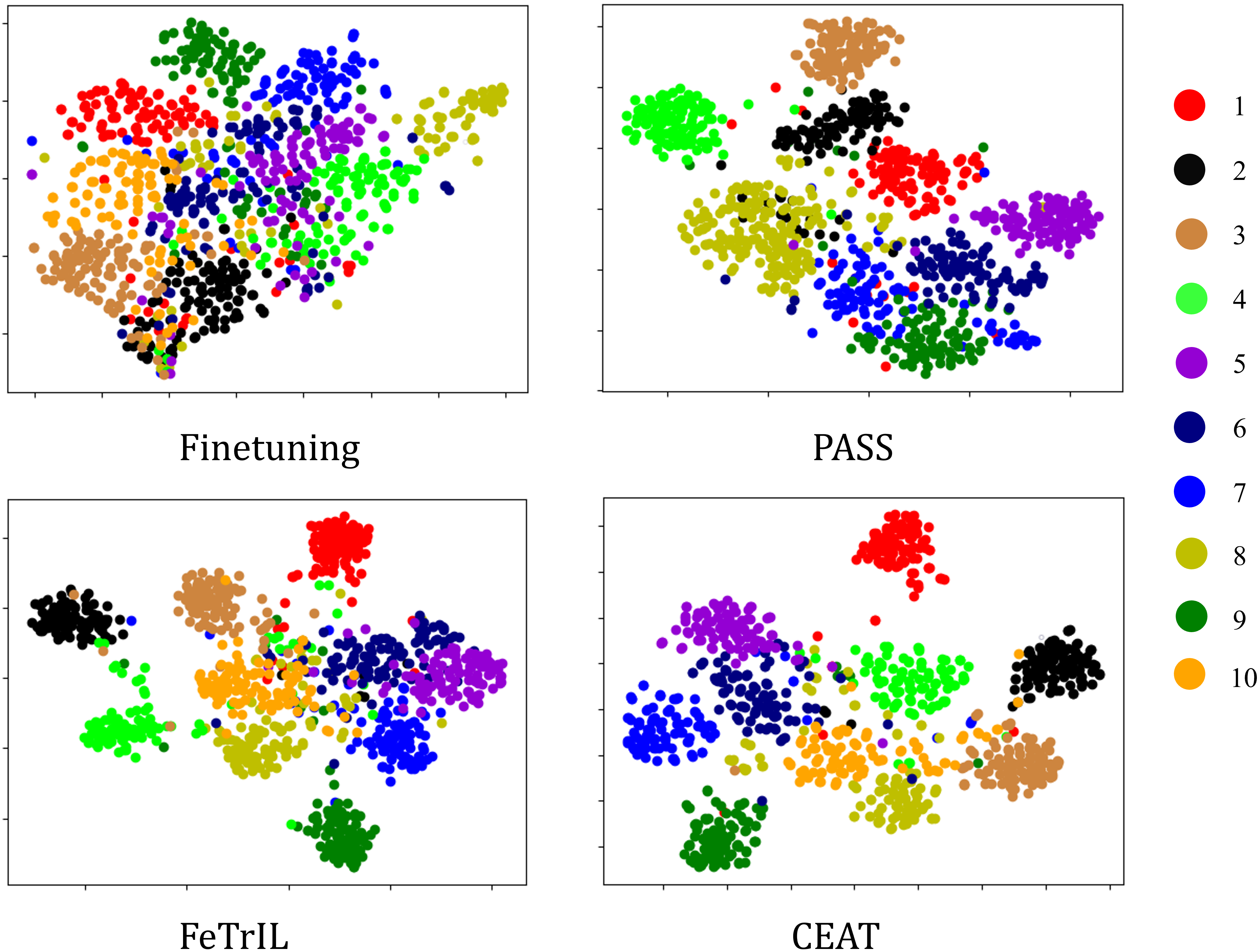}
\end{center}
\vspace{-0.2cm}
\caption{The visualization of the old feature using the final model on CIFAR-100 5 steps. We utilize the preceding images from the first 10 categories to evaluate the capability of the final model in preserving the distinguishing features of initial class categories.}
\label{fig:tsne}
\vspace{-0.2cm}
\end{figure}

\paragraph{Visualization.} To further illustrate the effectiveness of our approach, we show the visualization result with t-SNE\cite{van2008visualizing}. After the final task ends, we freeze the final model and use the previous images from the first 10 classes to compute the feature space distribution of these classes. The distribution of different classes is better distinguished, which means the anti-forgetting effect is better. We report the result in figure \ref{fig:tsne}. It is worth noting that the final model of our method can still distinguish the features of the first 10 classes well compared with previous methods\cite{zhu2021prototype,petit2023fetril}. This corresponds to our great anti-forgetting results in average forgetting.

\begin{table}[t]
\renewcommand{\arraystretch}{1.3}
  \renewcommand{\tabcolsep}{8.pt}
  \fontsize{12}{14}\selectfont
\centering
\resizebox{0.99\linewidth}{!}{
\begin{tabular}{lcccccc}
    \hline
          & \multicolumn{3}{c}{CIFAR-100}                                             & \multicolumn{3}{c}{TinyImageNet}                                           \\  \cline{2-7}
\multicolumn{1}{c}{\multirow{-2}{*}{\textbf{Methods}}}  & \multicolumn{1}{c}{5} & \multicolumn{1}{c}{10} & \multicolumn{1}{c}{20} & \multicolumn{1}{c}{5} & \multicolumn{1}{c}{10 } & \multicolumn{1}{c}{20} \\ 
    \hline
iCaRL-CNN\cite{ICARL} & \multicolumn{1}{c}{42.13}    & \multicolumn{1}{c}{45.69}     & 43.54     & \multicolumn{1}{c}{36.89}    & \multicolumn{1}{c}{36.70}     & 45.12     \\
iCaRL-NCM\cite{ICARL} & \multicolumn{1}{c}{24.90}    & \multicolumn{1}{c}{28.32}     & 35.53     & \multicolumn{1}{c}{27.15}    & \multicolumn{1}{c}{28.89}     & 37.40     \\
EEIL\cite{EEIL}      & \multicolumn{1}{c}{23.36}    & \multicolumn{1}{c}{26.65}     & 32.40     & \multicolumn{1}{c}{25.56}    & \multicolumn{1}{c}{25.91}     & 35.04     \\
UCIR\cite{lucir}     & \multicolumn{1}{c}{21.00}    & \multicolumn{1}{c}{25.12}     & 28.65     & \multicolumn{1}{c}{20.61}    & \multicolumn{1}{c}{22.25}     & 33.74     \\ \hline
LwF\_MC\cite{LWF}   & \multicolumn{1}{c}{44.23}    & \multicolumn{1}{c}{50.47}     & 55.46     & \multicolumn{1}{c}{54.26}    & \multicolumn{1}{c}{54.37}     & 63.54     \\
MUC\cite{liu2020more}       & \multicolumn{1}{c}{40.28}    & \multicolumn{1}{c}{47.56}     & 52.65     & \multicolumn{1}{c}{51.46}    & \multicolumn{1}{c}{50.21}     & 58.00     \\
PASS\cite{zhu2021prototype}      & \multicolumn{1}{c}{25.20}    & \multicolumn{1}{c}{30.25}     & 30.61     & \multicolumn{1}{c}{18.04}    & \multicolumn{1}{c}{23.11}     & 30.55     \\
SSRE\cite{zhu2022self}       & \multicolumn{1}{c}{18.37}    & \multicolumn{1}{c}{19.48}    & 19.00     & \multicolumn{1}{c}{9.17}    & \multicolumn{1}{c}{14.06}     & 14.20    \\ 
FeTrIL\cite{petit2023fetril}       & \multicolumn{1}{c}{14.46}    & \multicolumn{1}{c}{15.31}    & 17.08     & \multicolumn{1}{c}{8.02}    & \multicolumn{1}{c}{10.23}     & 10.64    \\ 
CEAT      & \multicolumn{1}{c}{\textbf{7.85}}    & \multicolumn{1}{c}{\textbf{8.04}}     & \textbf{8.68}     & \multicolumn{1}{c}{\textbf{6.10}}     & \multicolumn{1}{c}{\textbf{5.21}}     & \textbf{6.01}     \\ 
    \hline
\end{tabular}}
\caption{Results of average forgetting on 5, 10 and 20 phases in CIFAR100 and TinyImageNet. The results are obtained from \cite{zhu2022self}. Average forgetting demonstrates its ability to reduce catastrophic forgetting.}
\vspace{-0.2cm}
\label{tab:forget}
\end{table}

\begin{table}[ht]
  \renewcommand{\arraystretch}{1.2}
  \renewcommand{\tabcolsep}{10.pt}
  \centering
  \resizebox{0.9\linewidth}{!}{
    \begin{tabular}{l|l|ccc}
      \hline
      \multicolumn{1}{c|}{} & \multicolumn{1}{c|}{} & \multicolumn{3}{c}{\textbf{CIFAR-100}}    \\ \cline{3-5}  
      \multicolumn{1}{c|}{\multirow{-2}{*}{\textbf{Methods}}} & \multicolumn{1}{c|}{\multirow{-2}{*}{\textbf{Params}}} & \textit{P=20} & \textit{P=10} & \textit{P=5} \\
      \hline
      PASS-CNN &11.22M & 58.09  & 61.84 & 63.47\\
      PASS-ViT & 10.89M & 43.21 & 55.04 & 59.78  \\ 
      CEAT-ViT &10.89M  & 65.14 & 69.46 & 71.68   \\
      \hline
    \end{tabular}
  }
\caption{We reproduce PASS with the same ViT backbone, data augmentation, and hyperparameter tuning as our method on CIFAR-100. It is worth noticing that despite the adjustment of hyperparameters, PASS still performs disastrously based on ViT architecture, especially long-range increment learning.}
  \label{tab:vit}
\vspace{-0.2cm}
\end{table}

\subsection{Discussion on Motivation and Comparison}
\paragraph{Additional Motivation.} As we mentioned in related work, \cite{yu2021improving} have showcased that ViTs are more prone to forgetting when trained from scratch. Therefore, we have conducted the experiment to verify whether the previous method is suitable for the ViT structure. We reproduce the most representative method-PASS\cite{zhu2021prototype} with the same ViT backbone and data augmentation as our method. We tune all additional hyperparameters using 20\% of the training data as validation data like our model to reach a better performance. We report the results in table \ref{tab:vit}. It is worth noticing that despite the large adjustment of hyperparameters, the previous method still performs disastrously, especially the longer sequences of incremental training sessions. This motivates us to propose a new approach to apply the ViT architecture to the NECIL problem.

\begin{figure}[ht]
\begin{center}
    \includegraphics[width=0.4\textwidth]{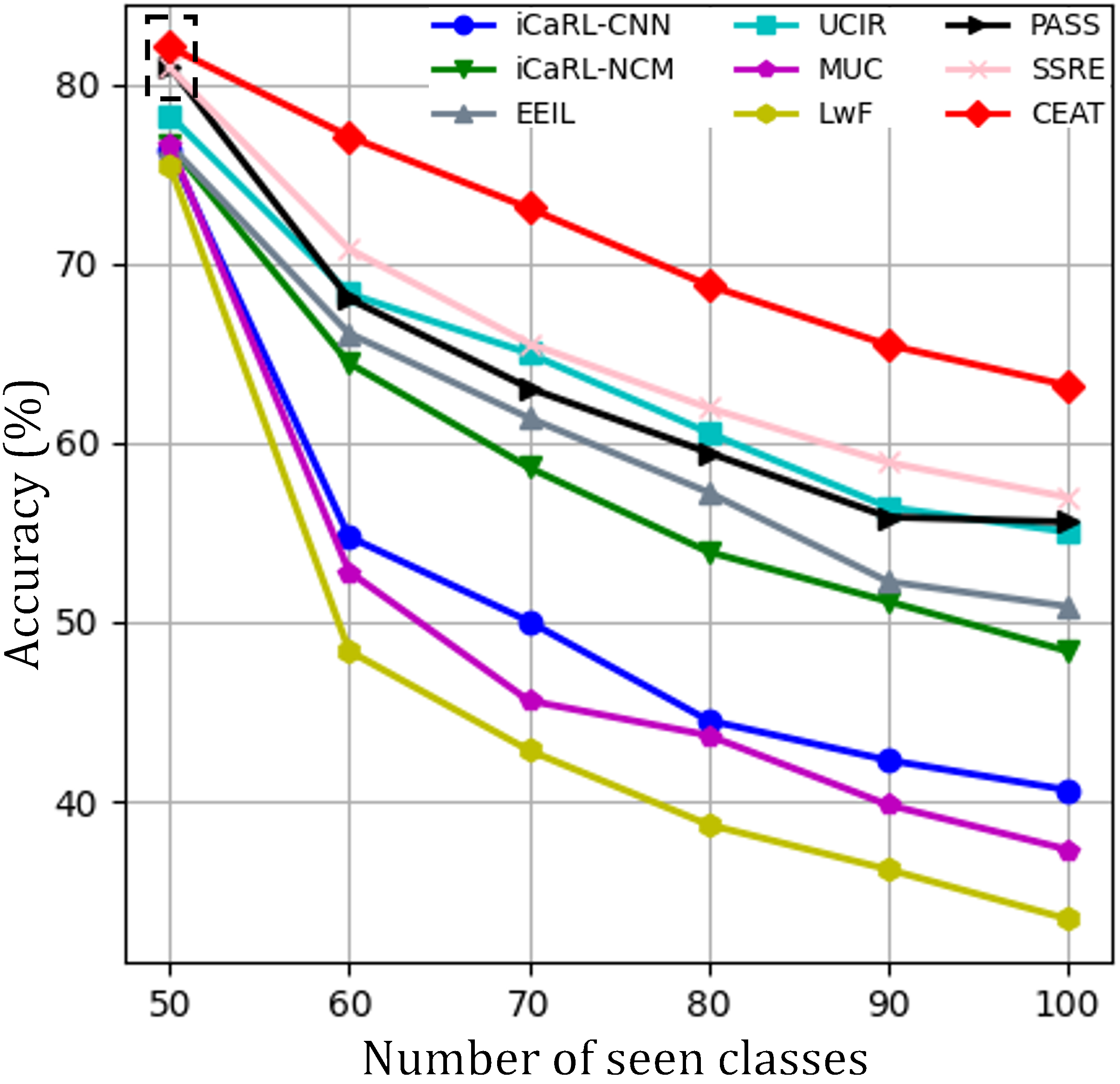}
\end{center}
\caption{The performance on CIFAR-100 5 steps. Note that at the first step before the continual process begins~(represented by a dotted rectangle), our model has performance comparable to other last methods~\cite{zhu2021prototype,zhu2022self}. This phenomenon illustrates that our method improves performance by solving the plasticity-stability dilemma.}
\vspace{-0.1cm}
\label{fig:cifar5}
\end{figure}

\paragraph{Comparison Analysis.} For a fair comparison, we compare our method with previous methods from multiple aspects. \textbf{Firstly}, in order to meet the limitation on the parameters of the NECIL problem, the final number of parameters of our model is 10.89M in CIFAR100 and 11.10M in ImageNet-Subset, slightly less than ResNet18~(11.22M) to ensure the fairness of the comparison. \textbf{Secondly}, we report the line graphs in figure~\ref{fig:cifar5}. It is worth noting that our method has a similar initial performance to other state-of-the-art methods and gets a better last accuracy. This phenomenon corresponds to our average forgetting result reported in the main text. \textbf{Lastly}, we report accuracy of other ViT-based general CIL method \cite{douillard2022dytox}, which stores 20 images per class in memory buffer. Our method still exhibits comparable performance, which shows that our method effectively fills the performance gap caused by the storage of examples, highlighting its practicality and effectiveness.

The above phenomenons proof of that our model can effectively apply ViT architecture to NECIL problems and reach a comparable performance with other ViT-based general CIL methods. The ablation study demonstrates how our contributions improve the model step-by-step.

\subsection{Ablation Analyses}
\paragraph{Contribution Ablation Analyses.} We show the contribution ablation result in table \ref{tab:ablation}. We established a baseline model consisting of six SABs and the Gaussian noise pseudo-features. \textbf{The baseline is the standard PASS model, which replaces ResNet18 with ViT}. To demonstrate the efficacy of our proposed approach, we incrementally incorporated components of Freeze, Batch Interpolation Pseudo-Features~(IPF), Continual Expansion and Absorption~(CEA), and Prototype Contrastive Loss~(PCL) into the baseline model. We conduct the experiments on three incremental settings of the CIFAR100 dataset. Firstly,
we freeze the backbone and tune the classifier with new data. Secondly, we replace Gaussian noise pseudo-features with our IPF, which has improved the performance significantly under the three settings. This result shows that the IPF effectively solves the problem of decision boundary degradation in previous work. In addition, The CEA 
expands the ex-fusion to learn new knowledge continuously while maintaining the previous parameters frozen, which makes impressive progress in different settings. The PCL enforces the separation of new and old classes and improves the learning of new classes. It's worth noting that the improvement of PCL in three settings is various. The more new classes that need to be learned in each phase, the more obvious the improvement of PCL becomes. This phenomenon may be caused that the model can learn well when facing fewer new classes so that the effect of PCL is not obvious. More ablation analyses about the inserted layer and inserted position are added in the Appendix.

\begin{table}[h]
  \renewcommand{\arraystretch}{1.2}
  \renewcommand{\tabcolsep}{8.pt}
  \centering
  \resizebox{0.99\linewidth}{!}{
    \begin{tabular}{ccccccc}
      \hline
      \multirow{2}{*}{Freeze} & \multirow{2}{*}{IPF} & \multirow{2}{*}{CEA} & \multirow{2}{*}{PCL} & \multicolumn{3}{c}{CIFAR-100}    \\
      \cline{5-7} 
      & & & & 20 phases & 10 phases & 5 phases \\ 
      \hline
      \ding{53}  & \ding{53} & \ding{53}  & \ding{53} & 43.21  & 55.04  & 59.78 \\
      $\surd$ & \ding{53} & \ding{53} & \ding{53} & 46.71 & 56.21 & 62.95 \\ 
      $\surd$ & $\surd$ & \ding{53} & \ding{53} & 50.99 & 61.15 & 65.06 \\
      $\surd$ & $\surd$ & $\surd$ & \ding{53} & 65.10 & 68.80 & 69.49  \\
      $\surd$ & $\surd$ & $\surd$ & $\surd$  & \textbf{65.14} & \textbf{69.46} & \textbf{71.68}  \\
      \hline
    \end{tabular}
  }
  \vspace{-0.1cm}
  \caption{Ablation study of our method on CIFAR-100 with different settings. We gradually add our contributions to the baseline model to prove its effectiveness.}
  \label{tab:ablation}
  \vspace{-0.2cm}
\end{table}

\section{Conclusion}
The objective of CIL is to continually acquire new classes while preventing the degradation of previously learned classes when they are not accessible. Experience-replay methods are to store a set of old samples to replay, which poses privacy and data leakage risks in the real world. To solve these problems, researchers propose the NECIL, which does not allow the model to store any old samples and does not use the pre-trained model. However, its performance is inferior to experience-replay methods due to the lack of old images. For these phenomena, our paper proposes a continual expansion and absorption transformer (CEAT), which meets the main challenges in NECIL and reaches state-of-the-art performance. Based on the unique ViT architecture, we employ ex-fusion layers to enable the learning of new knowledge while preserving previously acquired knowledge by fixing the main backbone. In order to improve the learning ability of new classes, we propose a Prototype Contrastive Loss to reduce the overlap of new and old classes. Moreover, to address the issue of classifier bias, we introduce a new batch interpolation pseudo-features approach. The ablation study provides further evidence of the efficacy of the proposed methods.

{\small
\bibliographystyle{ieee_fullname}
\bibliography{egbib}
}

\end{document}